\documentclass[11pt]{article}

\usepackage[final]{acl}
\usepackage{booktabs}
\usepackage{placeins}
\usepackage{amsmath}
\usepackage[table]{xcolor}

\usepackage{bm}
\usepackage{times}
\usepackage{latexsym}

\usepackage[T1]{fontenc}

\usepackage[utf8]{inputenc}

\usepackage{microtype}

\usepackage{inconsolata}
\usepackage{tabularx}
\newcolumntype{Y}{>{\centering\arraybackslash}X}

\usepackage{graphicx}
\usepackage{multicol}
\usepackage{multirow}
\usepackage{xcolor}
\usepackage{enumitem}
\usepackage{multirow}
\usepackage{alltt}
\usepackage[textsize=scriptsize]{todonotes}
\usepackage{amsfonts}


\title{Randomized YaRN Improves Length Generalization\\for Long-Context Reasoning}

\author{
Manas Mehta \quad  Fangcong Yin \quad  Greg Durrett \\
 New York University \\
  \texttt{\{mm14444,fy666,gdurrett\}@nyu.edu}
}

\begin{document}
\maketitle
\begin{abstract}
Large language models (LLMs) are typically pretrained on short sequences and then extended to work on longer sequences with additional training. However, such LLMs still struggle to further generalize to very long sequences. We propose Randomized YaRN, a training method that improves length generalization by combining YaRN-based positional extrapolation with randomized positional encoding and a length curriculum. During training on short context data, tokens are assigned YaRN positional encodings sampled from a larger position range, exposing the model to out-of-distribution positional representations even on short-context inputs. We evaluate Randomized YaRN on three challenging long-context reasoning benchmarks, BABILong, Multi-Round Coreference Resolution (MRCR), and LongBench v2. When training on data with short context, Randomized YaRN consistently improves reasoning performance on context lengths from 16K to 128K and outperforms standard fine-tuning, with the largest gains appearing at far out-of-distribution lengths. Our results suggest that progressively exposing models to OOD positional distributions provides an effective recipe for generalizable long-context reasoning.\footnote{Our code and data are available at \url{https://github.com/Manas-Mehta/Randomized-YaRN}.}
\end{abstract}

\begin{table*}[!t]

\centering

\small

\begin{tabular}{l rrrrr r >{\columncolor{gray!15}}r>{\columncolor{gray!15}}r>{\columncolor{gray!15}}r>{\columncolor{gray!15}}r>{\columncolor{gray!15}}r}

\toprule

\multirow{1}{*}{} & \multicolumn{6}{c}{ID} & \multicolumn{5}{c}{OOD} \\

 & 0K & 1K & 2K & 4K & 8K & \multicolumn{1}{c}{\textit{Avg.}} & \multicolumn{1}{>{\columncolor{gray!15}}r}{16K} & \multicolumn{1}{>{\columncolor{gray!15}}r}{32K} & \multicolumn{1}{>{\columncolor{gray!15}}r}{64K} & \multicolumn{1}{>{\columncolor{gray!15}}r}{\textbf{128K}} & \multicolumn{1}{>{\columncolor{gray!15}}r}{\textit{Avg.}} \\

\midrule


\multicolumn{12}{c}{\textit{Qwen2.5-7B-Instruct}} \\


\midrule

0-shot           & 20.3 & 28.8 & 22.4 & 23.9 & 21.0 & 23.3 & 20.3 & 23.9 & 19.7 & 18.7 & 20.7 \\

0-shot + YaRN    & 19.7 & 26.1 & 25.0 & 26.6 & 22.0 & 23.9 & 25.6 & 18.0 & 20.0 & 21.3 & 21.2 \\
\hline

LoRA                & \textbf{95.4} & 96.2 & \textbf{95.1} & \textbf{95.7} & 93.1 & \textbf{95.1} & 92.5 & 88.9 & 90.2 & 63.0 & 83.6 \\

Trained YaRN        & 87.2 & 91.7 & 92.4 & 91.5 & 89.5 & 90.5 & 89.5 & 86.2 & 82.3 & 67.2 & 81.3 \\

RPE                 & 79.3 & 83.3 & 82.9 & 85.6 & 83.6 & 82.9 & 84.9 & 82.0 & 79.0 & 72.8 & 79.7 \\

PoSE                & 91.2 & \textbf{96.6} & 93.4 & 91.5 & 89.5 & 92.4 & 87.2 & 79.7 & 75.4 & 70.2 & 78.1 \\

Randomized YaRN     & 89.5 & 90.2 & 90.8 & 92.5 & \textbf{93.8} & 91.4 & \textbf{93.8} & \textbf{93.4} & 90.2 & \textbf{83.9} & \textbf{90.3} \\

\midrule


\multicolumn{12}{c}{\textit{Olmo3-7B-Instruct}} \\


\midrule

0-shot           & 31.2 & 26.1 & 23.7 & 25.6 & 21.6 & 25.6 & 22.6 & 19.7 & 20.0 & 14.1 & 19.1 \\

0-shot + YaRN    & 31.2 & 26.5 & 23.4 & 26.2 & 21.6 & 25.8 & 22.0 & 22.0 & 19.3 & 16.7 & 20.0 \\
\hline

Trained YaRN    & 97.4 & \textbf{99.6} & 98.0 & 96.7 & 95.7 & 97.5 & \textbf{98.0} & 89.5 & 78.0 & 67.2 & 83.2 \\

PoSE            & \textbf{98.7} & 99.2 & \textbf{98.7} & \textbf{98.7} & 97.4 & \textbf{98.5} & 94.4 & 88.9 & 80.3 & 73.1 & 84.2 \\

Randomized YaRN & 96.4 & 98.9 & 95.1 & 98.0 & \textbf{98.7} & 97.4 & 96.4 & \textbf{91.5} & \textbf{87.2} & \textbf{76.7} & \textbf{88.0} \\

\midrule


\multicolumn{12}{c}{\textit{Qwen3-14B}} \\


\midrule

Trained YaRN    & \textbf{98.0} & \textbf{99.2} & 97.4 & \textbf{98.7} & 97.4 & \textbf{98.1} & 96.7 & 91.1 & 85.6 & 81.6 & 88.8 \\

PoSE            & 96.7 & 97.4 & 97.4 & 97.4 & 95.1 & 96.8 & 94.8 & 89.5 & 81.0 & 76.7 & 85.5 \\

Randomized YaRN & 93.1 & 97.7 & 96.4 & 97.0 & 97.4 & 96.3 & \textbf{97.4} & \textbf{95.1} & \textbf{90.5} & \textbf{87.5} & \textbf{92.6} \\

\bottomrule

\end{tabular}

\caption{In-distribution (ID) and out-of-distribution (OOD) evaluations on BABILong. Vanilla LoRA and RPE without YaRN during training are not supported for Olmo3-7B-Instruct as it is pre-trained with YaRN encodings, so they are excluded from evaluation; we also omit them for Qwen3-14B. Randomized YaRN outperforms all baselines on the OOD average across all three models, and particularly shows strength at the far OOD length of 128K.}

\label{tab:babilong_main}

\end{table*}

\section{Introduction}

Large language models (LLMs) are typically pre-trained on short sequences due to computational constraints. \emph{Length generalization}, the ability to extrapolate to input contexts longer than those seen during pre-training, has thus become a central challenge in long-context language modeling. A common approach to extending the context window of LLMs is to continue pre-training \citep{gao-etal-2025-train,grattafiori2024llama3herdmodels}
or fine-tune \citep{bai-etal-2024-longalign,lu2025controlled_study} on large-scale long-context corpora. However, this strategy primarily brings certain long sequences in-distribution through data engineering and computation, without necessarily imparting real length generalization. A separate line of work achieves generalization by modifying the positional encoding of LLMs to extend the context window during post-training \citep{ding2024longrope,chen2023extending}. While such methods perform well on simple long-context retrieval tasks such as Needle-in-a-Haystack, their effectiveness on \emph{reasoning} tasks at contexts beyond the training distribution remains mixed \citep{gelberg2026extending}.

\begin{figure}
\centering
\includegraphics[trim={0cm 3cm 5cm 0cm},clip,width=\columnwidth]{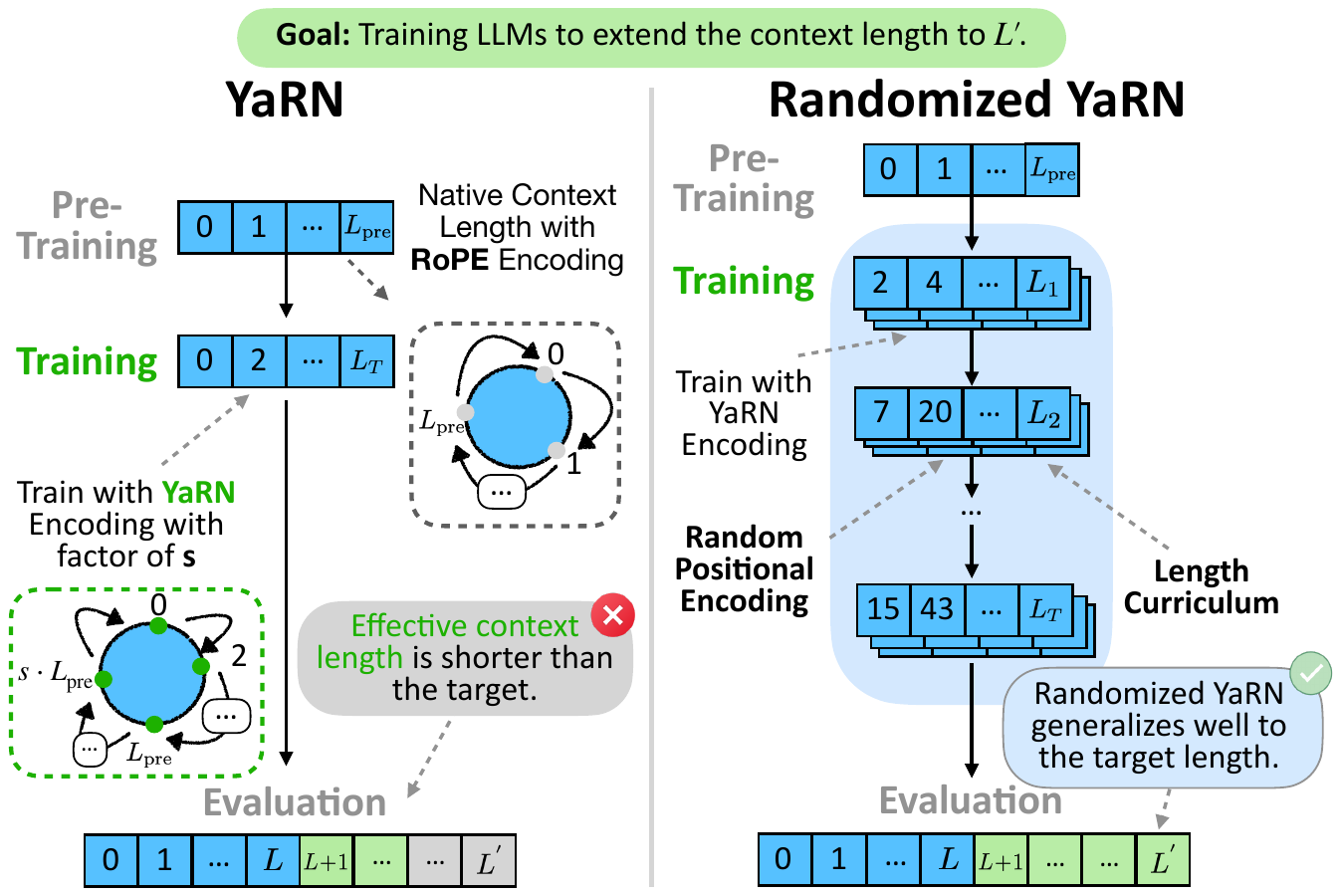}
\caption{Randomized YaRN trains LLMs for length generalization by: (1) Random position sampling: randomly sampling positions from a longer length distribution, assigning the corresponding YaRN encodings to each token in the short-context training data to expose models to OOD positional encodings; and (2) Length-generalization curriculum: gradually increasing the sampling length distribution as a training curriculum.} 
\label{figure:method_fig}
\end{figure}
 To improve length generalization for long-context reasoning, we propose \textbf{Randomized YaRN}, a long-context extension training method that builds on YaRN \citep{peng2023yarn}. Our approach draws on the idea of randomized positional encoding \citep{ruoss2023randomized, zhu2023pose}: as illustrated in Figure~\ref{figure:method_fig}, during training we assign each token in a short sequence the YaRN encoding vector of a position randomly sampled from a longer length distribution in sorted order, simulating out-of-distribution positions that standard training never exposes the model to. Crucially, the maximum length of this sampling distribution is increased gradually over the course of training to form a length-generalization curriculum.

We evaluate Randomized YaRN on three challenging long-context reasoning benchmarks, including BABILong \citep{kuratov2024babilong}, Multi-Round Coreference Resolution \citep{vodrahalli2024michelangelo}, and LongBench v2 \citep{bai2024longbench2}, and compare it against standard fine-tuning methods. We find that, given only limited short-context training data, modifying positional encodings yields substantially better length generalization than both inference-time methods and vanilla LoRA \citep{hu2022lora} fine-tuning. Moreover, models trained with Randomized YaRN on short-context data achieve stronger reasoning performance when evaluated at context lengths of 16K to 128K, consistently across different base models. Ablation studies further reveal that the length curriculum is important for robust generalization.

Our main contributions are as follows: (1) we propose a novel curriculum-based training method that improves the length generalization of LLMs; and (2) we present an effective recipe for context extension targeted at long-context reasoning tasks.

\section{Method}
\label{sec:method}

\subsection{Preliminaries}

\label{sec:yarn-prelim}
\textbf{Length Generalization.} A language model $M$ is pretrained on data with length $L_{\mathrm{pre}}$. During post-training, it is usually extended via further fine-tuning on a data distribution with a maximum context of $L$.  At test time, we evaluate it on tasks requiring even longer context length $L' > L$, and usually $L' > L_{pre}$.

\textbf{YaRN.} YaRN \citep{peng2023yarn} is a context extension method for models pretrained with Rotary Position Embeddings (RoPE) \citep{su2023roformerenhancedtransformerrotary}. YaRN defines a hyperparameter scale factor $s$ that can theoretically scale the context length from $L_{pre}$ up to $s\cdot L_{pre}$ by producing a modified positional encoding $\mathrm{YaRN}(m;\,s)$ for any position index $m$. At a high level, YaRN encoding leverages the NTK-by-parts interpolation for low-frequency RoPE dimensions to extend context length, and length-dependent attention temperature scaling to smooth the attention distribution at longer contexts. YaRN does not introduce any additional learnable parameters to models, and YaRN encoding has been widely adopted due to its flexibility \citep{qwen2025qwen25technicalreport}: it can be used in place of the original RoPE encoding at training \emph{and} inference time. 

\begin{table}
\footnotesize
\setlength{\tabcolsep}{0.5pt}
\begin{tabular}{lc>{\columncolor{gray!15}}c>{\columncolor{gray!15}}c>{\columncolor{gray!15}}c>{\columncolor{gray!15}}c>{\columncolor{gray!15}}c}
\toprule
\multirow{2}{*}{} & ID & \multicolumn{5}{c}{OOD} \\
\midrule
& 4-8K 
& \multicolumn{1}{>{\columncolor{gray!15}}l}{8-16K} 
& \multicolumn{1}{>{\columncolor{gray!15}}l}{16-32K} 
& \multicolumn{1}{>{\columncolor{gray!15}}l}{32-64K} 
& \multicolumn{1}{>{\columncolor{gray!15}}l}{64-128K} 
& \multicolumn{1}{>{\columncolor{gray!15}}l}{\textit{Avg.}} \\
\midrule

\multicolumn{7}{c}{\textit{Qwen2.5-7B-Instruct}} \\
\midrule

0-shot           & 38.9 & 36.5 & 46.5 & 16.5 &  5.6 & 26.3 \\
0-shot + YaRN    & 34.6 & 30.2 & 31.9 & 24.2 & 11.4 & 24.4 \\
\hline
LoRA                & \textbf{99.8} & \textbf{96.6} & 74.6 & 54.5 & 31.7 & 64.4 \\
Trained YaRN        & 89.1          & 69.2          & 61.9 & 61.9 & 44.1 & 59.3 \\
RPE                 & 81.7          & 59.2          & 67.7 & 64.6 & 52.8 & 61.1 \\
PoSE                & 77.8          & 54.5          & 43.1 & 53.6 & 37.5 & 47.2 \\
Randomized YaRN     & 96.0          & 69.2          & \textbf{79.8} & \textbf{72.8} & \textbf{68.8} & \textbf{72.7} \\

\midrule

\multicolumn{7}{c}{\textit{Olmo3-7B-Instruct}} \\
\midrule

0-shot           & 21.3 &  8.7 & 10.7 &  8.7 &  1.0 &  7.3 \\
0-shot + YaRN    & 21.3 &  7.5 & 12.0 &  9.7 &  6.5 &  8.9 \\
\hline
Trained YaRN        & 72.4          & \textbf{63.9} & 40.5 & 36.5 & 11.9 & 38.2 \\
PoSE                & 65.8          & 56.9          & 35.3 & 32.2 &  9.3 & 33.4 \\
Randomized YaRN     & \textbf{77.4} & 62.8          & \textbf{53.1} & \textbf{39.7} & \textbf{19.5} & \textbf{43.8} \\

\midrule

\multicolumn{7}{c}{\textit{Qwen3-14B}} \\
\midrule

Trained YaRN        & \textbf{96.3} & 93.6          & 84.3 & 80.2 & 55.1 & 78.3 \\
PoSE                & 89.3          & 70.7          & 64.2 & 57.6 & 41.2 & 58.4 \\
Randomized YaRN     & 95.9          & \textbf{99.8} & \textbf{93.6} & \textbf{90.2} & \textbf{81.9} & \textbf{91.4} \\

\bottomrule
\end{tabular}
\caption{In-distribution (ID) and out-of-distribution (OOD) evaluations on MRCR. \emph{Avg.} indicates average accuracy over OOD evaluations. Randomized YaRN outperforms baselines at OOD context length. }
\label{tab:mrcr_main}
\end{table}

\subsection{Randomized YaRN with a Length Curriculum}
\label{sec:randomized-yarn}

Our method, Randomized YaRN, improves upon YaRN for better length generalization. During training on data of length $L$, we apply standard LoRA fine-tuning to model parameters in all linear layers of $M$, and use YaRN encodings throughout training with the scale factor $s$ as a hyperparameter. At the same time, we apply the randomized position encoding sampling scheme on top of YaRN encodings, with a curriculum that gradually grows the sampling range over epochs. 

\textbf{Randomized Positional Encoding (RPE).} First introduced by \citet{ruoss2023randomized}, this training strategy exposes a model to OOD position indices via randomly assigning positional encodings of another position from a length distribution much larger than the maximum training length. 

Following the notation in \citet{ruoss2023randomized}, let $P_{k} = \{S \subseteq \{1,\dots,L'\} \mid |S| = k\}$ be the set of $k$-subsets of $\{1,\dots,L'\}$. For each training batch, RPE
samples a subset of indices without replacement, then sorts them in ascending order to produce $I \;\sim\; \mathcal{U}(P_L),  I = \{i_1 < i_2 < \dots < i_L\}$. RPE replaces the standard positional encoding of the $j$-th token with the encoding evaluated at the sampled index: $
\mathrm{RPE}(j) \;:=\; \mathrm{PE}(i_j)$ where $1\le j\le L.$
Because $I$ is sorted, the token order is preserved, but the absolute indices fed into the positional encoding span a much larger range (up to $L'$) than the actual sequence (up to $L$). This strategy trains LLMs to observe encodings outside the training length distribution, and has been adapted by other context extension methods \citep{zhu2023pose}.

\textbf{Position sampling.} Our method applies randomized position sampling to YaRN encodings during training as follows. We train $M$ for $T$ epochs on the data distribution of length $L$. Let $L_t$ denote the maximum sampling length used at epoch $t$, and $L_{t} < L_{T}$ for $t < T$. At each training step in epoch $t$, we draw a set of positions
\begin{gather}
I^{(t)} \;\sim\; \mathcal{U}(P_L^{L_t}) \\
P_L^{L_t} = \bigl\{S\subseteq\{1,\dots,L_t\} \,\big|\, |S|=L\bigr\}
\end{gather}

We sort the set in ascending order as $I^{(t)}=\{i_1<\dots<i_L\}$, and assign the $j$-th token of the sequence the YaRN encoding of the sampled position, $\widetilde{\mathrm{PE}}(j) \;:=\; \mathrm{YaRN}\bigl(i_j;\,s\bigr)$, where the YaRN scale $s$ is a hyperparameter. 
YaRN is able to scale up to $s\cdot L_{\mathrm{pre}}$, but the observed rotations themselves are still not seen by the model during training, even though they are in the range of rotation lengths that can be observed. Randomizing position sampling enables the model to observe such rotations that it might encounter during evaluation in longer contexts. 

\textbf{Length-generalization curriculum.} Instead of directly setting $L_t$ to the target length $L'$ from the start, we use a curriculum in which $L_t$ is gradually increased across $T$ training epochs: $L_1 \leq L_2 \leq \dots \leq L_T$, e.g., $L_t \in \{8\text{K}, 16\text{K}, 24\text{K},\dots\}$. 

Intuitively, the curriculum enables the model to first learn to use extrapolated YaRN encodings on a moderate position range, and gradually generalize to the larger out-of-distribution indices that the final target length requires.

\textbf{Inference.} At inference time we use standard YaRN, i.e., the $j$-th token is embedded as $\mathrm{YaRN}(j;s')$ with inference scale $s'$. Usually $s'$ is the same as the training scale $s$, but we show that  $s' > s$ can unlock generalization even beyond $s\cdot L_{\mathrm{pre}}$; details are in Appendix ~\ref{appx:baseline_imp}. Randomized position sampling is applied only during training.

\begin{table}[t]
\footnotesize
\setlength{\tabcolsep}{4pt}
\begin{tabularx}{\columnwidth}{l>{\columncolor{gray!15}}Y>{\columncolor{gray!15}}Y>{\columncolor{gray!15}}Y}
\toprule
\multirow{2}{*}{} & \multicolumn{3}{c}{OOD} \\
\midrule
& \multicolumn{1}{>{\columncolor{gray!15}}c}{32-64K} 
& \multicolumn{1}{>{\columncolor{gray!15}}c}{64-128K} 
& \multicolumn{1}{>{\columncolor{gray!15}}c}{\textit{Avg.}} \\
\midrule

\multicolumn{4}{c}{\textit{Qwen2.5-7B-Instruct}} \\
\midrule

0-shot           & 34.9 & 32.3 & 33.6 \\
0-shot + YaRN    & 37.9 & 30.8 & 34.3 \\
\hline
Trained YaRN        & 42.4          & 36.9          & 39.7 \\
Randomized YaRN     & \textbf{43.9} & \textbf{38.5} & \textbf{41.2} \\

\midrule

\multicolumn{4}{c}{\textit{Olmo3-7B-Instruct}} \\
\midrule

0-shot           & 28.2 & 19.4 & 23.8 \\
0-shot + YaRN    & 31.0 & 19.4 & 25.2 \\
\hline
Trained YaRN        & 29.6          & 35.8          & 32.7 \\
Randomized YaRN     & \textbf{32.4} & \textbf{41.8} & \textbf{37.1} \\

\bottomrule
\end{tabularx}
\caption{Out-of-distribution (OOD) evaluations on LongBench v2. All models are fine-tuned on the $\leq$32K bin, so both evaluation bins are OOD. \emph{Avg.} indicates average accuracy over OOD evaluations. Randomized YaRN is the best method on both models, improving over Trained YaRN by 4.4 points on Olmo3-7B and 1.5 points on Qwen2.5-7B.}
\label{tab:longbench_main}
\end{table}

\begin{table}[t]
\footnotesize
\setlength{\tabcolsep}{0.5pt}
\begin{tabular}{lc>{\columncolor{gray!15}}c>{\columncolor{gray!15}}c>{\columncolor{gray!15}}c>{\columncolor{gray!15}}c>{\columncolor{gray!15}}c}
\toprule
                 & \multicolumn{1}{c}{ID}   & \multicolumn{5}{c}{OOD}                                                                                                                      \\
                 \midrule 
                 & \multicolumn{1}{c}{4-8K} & \multicolumn{1}{>{\columncolor{gray!15}}c}{8-16K} & \multicolumn{1}{>{\columncolor{gray!15}}c}{16-32K} & \multicolumn{1}{>{\columncolor{gray!15}}c}{32-64K} & \multicolumn{1}{>{\columncolor{gray!15}}c}{64-128K} & \multicolumn{1}{>{\columncolor{gray!15}}c}{\textit{Avg.}} \\
                 \midrule
RPE              & \textbf{81.7}   & \textbf{59.2}             & \textbf{67.7}              & \textbf{64.6}              & 52.8                        & \textbf{61.1}            \\
- w/o Curriculum & 53.1            & 49.9                      & 64.1                       & 59.0                        & \textbf{55.1}               & 57.0                     \\
                 \midrule
Randomized YaRN  & \textbf{96.0}            & \textbf{69.2}             & \textbf{79.8}              & 72.8                        & \textbf{68.8}               & \textbf{72.7}            \\
- w/o Curriculum & 81.3                     & 65.5                      & 66.6                       & \textbf{74.9}               & 58.8                        & 66.5                     \\

\bottomrule
\end{tabular}
\caption{Ablation of the length curriculum for RPE and Randomized YaRN on Qwen2.5-7B-Instruct for MRCR. All rows use YaRN at inference. \emph{Avg.} indicates average accuracy over OOD evaluations. Both training methods degrade without the length curriculum.}
\label{tab:ablation}
\end{table}

\section{Experimental Setup}
\textbf{Tasks.} We evaluate Randomized YaRN on three representative challenging long-context reasoning tasks. (1) \textbf{BABILong} \citep{kuratov2024babilong} evaluates multi-hop reasoning capability over multiple needles in a long passage. We focus on the challenging three-hop reasoning subset to evaluate complex long-context reasoning; (2) \textbf{Multi-Round Coreference Resolution (MRCR)} \citep{vodrahalli2024michelangelo} is a long-context benchmark that requires reasoning-based retrieval in multi-turn conversations; (3) \textbf{LongBench v2} \citep{bai2024longbench2} is a long-context benchmark of four-way multiple-choice questions over naturally occurring documents with multiple task categories including QA, long in-context learning, and code understanding. For evaluation, we use exact match for BABILong, a similarity-based metric for MRCR, and answer accuracy for LongBench v2, following \citet{ye2026dyscodynamicattentionscalingdecoding}. More details can be found in Appendix~\ref{appx:task_details}.

\textbf{Data.} We evaluate length generalization through small-scale (<5K training examples) fine-tuning on short context data and evaluate models on out-of-distribution (OOD) examples, where \emph{OOD means that the inference-time context length is longer than the fine-tuning context length}. The context ranges are dataset-specific as the following: (1) For BABILong, we train on 4K examples from 0--8K and evaluate on a challenging subset of up to 305 examples per length bin;  (2) For MRCR, we train on 60 examples of 4--8K context, and evaluate on 147 examples;  (3) LongBench v2 provides no short-context split, so we instead fine-tune on all 116 examples of $\leq$32K context and evaluate on the 32--64K and 64--128K bins, giving a purely OOD evaluation. More details are in Appendix~\ref{appx:task_details}.

\textbf{Models and Training.} We train on Qwen2.5-7B-Instruct \citep{qwen2025qwen25technicalreport} and Olmo3-7B-Instruct \citep{olmo2026olmo3} that use different attention architectures and natively support 32K and 64K context, respectively. To test whether our method holds at larger scale, we train Qwen3-14B \citep{yang2025qwen3technicalreport}, which natively supports 40K context, on BABILong and MRCR; we disable its thinking mode so that all models are evaluated under the same direct-answer protocol. We use LoRA for all training methods. Our method involves two key hyperparameters, the maximum length of the curriculum $L_{T}$ and the YaRN factors during training ($s$) and inference ($s'$); we discuss details in Appendix~\ref{appx:training_details}.

\textbf{Baselines.} We compare our method with the following baselines. 
(1) \textbf{Zero-shot} evaluation with or without YaRN at inference time; 
(2) \textbf{LoRA}: Vanilla LoRA fine-tuning without YaRN encoding during training and inference;  (3) \textbf{Trained YaRN}: LoRA fine-tuning with YaRN during training and inference; (4) \textbf{RPE}: Training with randomized position encoding without YaRN during training, but with YaRN at inference; (5) \textbf{PoSE} \citep{zhu2023pose}: Training with positional skip-wise encoding, a modified version of RPE that splits each input into two chunks and inserts a random offset between them to simulate positions up to a fixed target length, with YaRN at inference. Because Olmo3 is pre-trained with YaRN encodings, vanilla LoRA and RPE without YaRN during training are not applicable, so we do not include them in evaluation. Configuration details are in Appendix~\ref{appx:training_details}.

\section{Results}
\label{sec:results}

\textbf{Zero-shot inference with YaRN is limited.} A common way for applying YaRN is to only use YaRN encoding at inference time instead of during training\citep{qwen2025qwen25technicalreport}, but we find it limited for OOD length generalization.
Table ~\ref{tab:babilong_main} shows that on BABILong, adding YaRN at inference time only improves the average OOD accuracy by 1\%, and similarly for MRCR in Table~\ref{tab:mrcr_main} and LongBench v2 in Table~\ref{tab:longbench_main}. These results show that YaRN at inference time is limited, and training is necessary for better length generalization.

\textbf{Randomized YaRN outperforms baselines on OOD context length.} Table~\ref{tab:babilong_main} shows that on BABILong, Randomized YaRN achieves 90.3\% accuracy with Qwen2.5-7B, 88.0\% on Olmo3-7B, and 92.6\% on Qwen3-14B when evaluated on OOD test examples with context length ranging from 16K to 128K, beating all baselines on average. Similarly, on MRCR (Table~\ref{tab:mrcr_main}), Randomized YaRN outperforms all baselines by a large margin on OOD length for all three models. The same holds on LongBench v2 (Table~\ref{tab:longbench_main}), where Randomized YaRN achieves 41.2\% with Qwen2.5-7B and 37.1\% on Olmo3-7B on OOD examples with context length ranging from 32K to 128K. Randomized YaRN performs particularly well on examples in the far OOD length distribution (128K for BABILong and 64--128K bin for MRCR). 

These results highlight the strength of our method for improving length generalization in long-context reasoning.

\textbf{Our length curriculum is critical for length generalization.} Table~\ref{tab:ablation} shows the ablation results for RPE and Randomized YaRN on Qwen2.5-7B-Instruct for MRCR. For cleaner ablations, we perform YaRN at inference time for all results. Without the curriculum, OOD performance substantially decreases for both methods by 4.1 \% for RPE and 6.2 \% for Randomized YaRN, showing the importance of gradually increasing the distribution of the modified positional encodings during training.

\section{Related Work}

\textbf{Length generalization for long-context reasoning.} Length generalization has been a challenging task for LLMs \citep{anil2022exploring}, partially due to the limitations of the transformer architecture \citep{zhou2024transformers}. This limits the pretrained LLMs that mainly observe short sequences during pre-training in long-context reasoning \citep{yen2025helmet,ye2025longproc,bai-etal-2024-longbench}. 

\textbf{Context extension methods for LLMs.} Context extension methods extend LLMs beyond the pretrained context length. Data-centric approaches expose LLMs to longer distributions via extensive post-training on long-context instruction following data \citep{gao-etal-2025-train,bai-etal-2024-longalign} or synthetic long-context corpora \citep{zhao2024longskyworktrainingrecipeefficiently,zhao2025understanding}. Other approaches focus on modifying transformer modules, including attention \citep{ye2026dyscodynamicattentionscalingdecoding,zhang-etal-2025-query-focused} and positional encoding \citep{ding2024longrope,ruoss2023randomized}, for better long context performance. While inference-time extrapolation of positional encodings \citep{chen2023extending,zhang2024found,xu-etal-2025-extending} is used in the literature, training with the modified positional encodings has been shown more effective \citep{wu2024an,peng2023yarn}.

\section{Conclusion}

We introduce Randomized YaRN, a training method for improving length generalization in LLMs by enhancing YaRN encodings with random position sampling and length curriculum during training. Across long-context reasoning benchmarks, we show that Randomized YaRN improves performance at out-of-distribution context lengths compared with other training baselines, and highlight the importance of length-based training curriculum for robust length generalization. 

\section*{Limitations}

Our work has the following limitations: (1) Our method mainly applies to scenarios where the inference context length during evaluation is much longer than that during training. (2) Due to computational constraints, our experiments only evaluate 7B and 14B-scale models; (3) We only evaluate long-context reasoning on English-only data.

\section*{Acknowledgments}
We thank Xi Ye for helpful feedback on a draft of this work.  This work was supported by NSF CAREER Award IIS-2145280, NSF grant IIS-2433071, the NSF AI Institute for Foundations of Machine Learning (IFML), and the NSF under Cooperative Agreement 2421782 and the Simons Foundation grant MPS-AI-00010515 awarded to the NSF-Simons AI Institute for Cosmic Origins — CosmicAI, \url{https://www.cosmicai.org/}. This work was also partially supported by the Sloan Foundation. Finally, this work has been supported by a compute grant from NVIDIA.

\bibliography{custom}

\appendix

\begin{table*}
\centering
\small
\setlength{\tabcolsep}{5pt}
\begin{tabular}{l l c c l c}
\toprule
\textbf{Dataset} & \textbf{Method} & \textbf{$s$} & \textbf{$s'$} & \textbf{$L_t$ Curriculum} & \textbf{\# Epoch} \\
\midrule
\midrule
\multicolumn{6}{c}{\textit{Qwen2.5-7B-Instruct}} \\
\midrule
\multirow{5}{*}{BABILong}
  & LoRA            & ---  & ---  & ---                                        & 2  \\
  & Trained YaRN    & 2    & 4    & ---                                        & 2  \\
  & RPE             & ---  & 4    & 8K $\to$ 16K                               & 2  \\
  & PoSE            & ---  & 4    & 32K (fixed)                                & 2  \\
  & Randomized YaRN & 2    & 4    & 8K $\to$ 16K                               & 2  \\
\cmidrule(lr){1-6}
\multirow{5}{*}{MRCR}
  & LoRA            & ---  & ---  & ---                                        & 5  \\
  & Trained YaRN    & 2    & 4    & ---                                        & 5  \\
  & RPE             & ---  & 4    & 8K $\to$ 10K $\to$ 12K $\to$ 16K (ep4--5)  & 5  \\
  & PoSE            & ---  & 4    & 32K (fixed)                                & 5  \\
  & Randomized YaRN & 2    & 4    & 8K $\to$ 10K $\to$ 12K $\to$ 16K (ep4--5)  & 5  \\
\cmidrule(lr){1-6}
\multirow{2}{*}{LongBench v2}
  & Trained YaRN    & 2    & 4    & ---                                        & 5  \\
  & Randomized YaRN & 2    & 4    & 32K $\to$ 40K $\to$ 48K $\to$ 64K (ep4--5) & 5  \\
\midrule
\midrule
\multicolumn{6}{c}{\textit{Olmo3-7B-Instruct}} \\
\midrule
\multirow{3}{*}{BABILong}
  & Trained YaRN    & 16   & 16   & ---                                        & 5  \\
  & PoSE            & 16   & 16   & 32K (fixed)                                & 5  \\
  & Randomized YaRN & 16   & 16   & 8K $\to$ 16K $\to$ 32K (ep3--5)            & 5  \\
\cmidrule(lr){1-6}
\multirow{3}{*}{MRCR}
  & Trained YaRN    & 16   & 16   & ---                                        & 10 \\
  & PoSE            & 16   & 16   & 32K (fixed)                                & 10 \\
  & Randomized YaRN & 16   & 16   & 8K (ep1--2) $\to$ 16K $\to$ 24K $\to$ 32K (ep5--10) & 10 \\
\cmidrule(lr){1-6}
\multirow{2}{*}{LongBench v2}
  & Trained YaRN    & 16   & 16   & ---                                        & 5  \\
  & Randomized YaRN & 16   & 16   & 32K $\to$ 64K $\to$ 128K (ep3--5)          & 5  \\
\midrule
\midrule
\multicolumn{6}{c}{\textit{Qwen3-14B}} \\
\midrule
\multirow{3}{*}{BABILong}
  & Trained YaRN    & 2    & 4    & ---                                        & 2  \\
  & PoSE            & ---  & 4    & 32K (fixed)                                & 2  \\
  & Randomized YaRN & 2    & 4    & 8K $\to$ 16K                               & 2  \\
\cmidrule(lr){1-6}
\multirow{3}{*}{MRCR}
  & Trained YaRN    & 2    & 4    & ---                                        & 5  \\
  & PoSE            & ---  & 4    & 32K (fixed)                                & 5  \\
  & Randomized YaRN & 2    & 4    & 8K $\to$ 10K $\to$ 12K $\to$ 16K (ep4--5)  & 5  \\
\bottomrule
\end{tabular}
\caption{Per-method configuration of YaRN, RPE, and PoSE hyperparameters. $s$ is the YaRN scale factor during training and $s'$ is the factor during inference. The length curriculum changes the maximum sampling length $L_t$ at every epoch unless a range is given in parentheses, which indicates the epochs over which $L_t$ is held constant. PoSE uses no curriculum, so the corresponding column reports its fixed target length. ``---'' indicates the corresponding hyperparameter is not applicable to the setting. }
\label{tab:method-config}
\end{table*}

\section{Details of Task Setup}
\label{appx:task_details}

\subsection{MRCR}

MRCR \citep{vodrahalli2024michelangelo} is a long-context reasoning dataset that requires retrieving information from long passages by understanding latent structures. In MRCR, each example consists of a long multi-turn conversation containing a sequence of assistant responses corresponding to different user messages. Given a query, the model must locate the corresponding assistant response, reproduce it verbatim, and then perform an additional action based on the retrieved information. For example, the input context may contain two ``formal letters about planning'' or four ``short essays about knowledge'', and the query may be \emph{``prepend [random string] to the 1st formal letter about planning''}. The model must locate the correct response, reproduce it verbatim, and prepend the specified random string. In addition, each example contains 2, 4, or 8 distractor responses that closely resemble the target response, making the task a challenging benchmark for multi-hop reasoning over long contexts.

\paragraph{Data Statistics}

MRCR is a relatively small dataset with limited training and evaluation data available. We train on 60 examples from the 4–8K context-length bin and evaluate on all 147 held-out examples across five bins (4–8K, 8–16K, 16–32K, 32–64K, and 64–128K), with 26, 30, 30, 30, and 31 examples per bin respectively.

\paragraph{Evaluation Metrics}

We use the same similarity-based metric as the official MRCR benchmark implementation. The score is computed as the character-level similarity between the model output and the target response if the required action (e.g., prepending a prefix) is performed correctly; otherwise, if the random prefix is missing, the score is set to 0.

\subsection{BABILong}

BABILong \citep{kuratov2024babilong} is a long-context reasoning benchmark built upon BaBI \citep{weston2015aicompletequestionansweringset} and consists of multiple reasoning subtasks. We focus on a challenging subtask (\emph{QA3}) that requires multi-hop reasoning with 3 supporting facts and has not shown saturated performance even among frontier LLMs. In this subtask, each example is constructed by embedding three bAbI facts into long passages of irrelevant text drawn from PG19, a corpus of English-language books, to pad the context to the target length. The model must accurately locate the relevant facts and reason over them to answer a question.

To stress-test length generalization in long-context reasoning, we focus on a smaller and more challenging subset involving multi-entry queries. Each example contains a state-tracking task where an object is moved across locations by different people, and the model must determine the object's final location. For example, one question may ask \emph{``where was the football before the garden?''}. The football is moved into the garden twice: first from the bedroom and later from the bathroom. The correct answer is \emph{bathroom} (the most recent transition), while \emph{bedroom} is also a plausible candidate. Therefore, the model must correctly reason about temporal order rather than simply retrieve a single mention.

\begin{table*}[!t]
\centering
\footnotesize
\setlength{\tabcolsep}{3pt}
\begin{tabular}{l l c c c c}
\toprule
\textbf{Model} & \textbf{Dataset} & \textbf{LR} & \textbf{Batch Size} & \textbf{Warmup Ratio} & \textbf{Max Length} \\
\midrule
Qwen2.5        & BABILong     & 5e-5 & 4 & 0.15 & 9{,}216  \\
Qwen2.5        & MRCR         & 2e-4 & 4 & 0.10 & 8{,}192  \\
Qwen2.5        & LongBench v2 & 5e-5 & 4 & 0.10 & 32{,}768 \\
Olmo3          & BABILong     & 1e-4 & 4 & 0.10 & 9{,}216  \\
Olmo3          & MRCR         & 2e-4 & 4 & 0.10 & 8{,}192  \\
Olmo3          & LongBench v2 & 5e-5 & 4 & 0.10 & 32{,}768 \\
Qwen3-14B      & BABILong     & 5e-5 & 4 & 0.15 & 9{,}216  \\
Qwen3-14B      & MRCR         & 2e-4 & 4 & 0.10 & 8{,}192  \\
\bottomrule
\end{tabular}
\caption{LoRA training hyperparameters for all training methods.
}
\label{tab:hparams}
\end{table*}

\paragraph{Data Statistics}
We train on 4K total training examples with 1K examples from four context-length bins (0K, 2K, 4K, and 8K). The 0K bin contains only the bAbI fact sequence without any padding passages. For evaluation, we use a 305-sample per bin subset across all nine context-length bins (0K, 1K, 2K, 4K, 8K, 16K, 32K, 64K, and 128K), except 264 samples at 1K and 304 at 2K.

\paragraph{Evaluation Metrics}

Following the official BABILong implementation, we use exact match against the ground-truth answer as the evaluation metric.

\subsection{LongBench v2}

LongBench v2 \citep{bai2024longbench2} is a long-context benchmark of four-way multiple-choice questions over naturally occurring documents, spanning six task categories: single-document QA, multi-document QA, long in-context learning, long-dialogue history understanding, code repository understanding, and long structured data understanding. Each example presents a document, a question, and four answer options, and the model must select the correct option. Unlike MRCR and BABILong, whose contexts are assembled synthetically to reach a target length, LongBench v2 contexts are real documents, so context length and task difficulty are not independently controlled.

Because every example in LongBench v2 already exceeds 8K tokens, the dataset provides no short-context split to fine-tune on. We therefore treat the shortest available contexts as the training distribution: all examples of at most 32K tokens are used for training, and evaluation is performed only on longer contexts, so that every evaluation example is out-of-distribution with respect to the fine-tuning length. Examples longer than 128K tokens are excluded, as they exceed the evaluation context window. To ensure that gains reflect length generalization rather than transfer to unseen task types, we retain an evaluation sub-domain only if it appears in at least three training examples, which removes sub-domains that are absent or nearly absent from the training split.

\paragraph{Data Statistics}

Context lengths are measured with each model's own tokenizer, so the split differs slightly across models. For Qwen2.5-7B-Instruct we train on 116 examples and evaluate on 131, with 66 and 65 examples in the 32--64K and 64--128K bins respectively. For Olmo3-7B-Instruct we train on 119 examples and evaluate on 138, with 71 and 67 examples in the two bins. Because the evaluation sets are model-specific, we compare methods within a model and do not compare across models.

\paragraph{Evaluation Metrics}

We use the official LongBench v2 zero-shot prompt template, which presents the document, the question, and the four options, and asks the model to respond in the form ``The correct answer is (X)''. Models answer directly rather than producing a chain of thought. The metric is accuracy over the four answer options.

\section{Training Details}
\label{appx:training_details}

\subsection{Base Models}

We use Qwen2.5-7B-Instruct, Olmo3-7B-Instruct, and Qwen3-14B as the base models for evaluating our method and baselines.

\paragraph{Attention Mechanism}

Qwen2.5 and Qwen3 both use grouped query attention \citep{ainslie2023gqa}, an attention mechanism that attends over the entire context window. In contrast, Olmo3 uses sliding-window attention in 24 of its 32 layers, with full attention in the remaining 8 layers. Consistent with the pretraining process of Olmo3, we apply YaRN and RPE positional encoding modifications exclusively to the full-attention layers; sliding-window layers retain vanilla RoPE throughout. LoRA parameters are applied uniformly to all layers.

\paragraph{Native YaRN Support}

Olmo3 is pre-trained with YaRN built in using a scale factor of $s{=}8$, enabling a native context length of up to $64$K, whereas Qwen2.5 and Qwen3 do not include native YaRN support. As a result, RPE training without YaRN is not an applicable baseline for Olmo3 and is therefore excluded from our evaluation.

\subsection{LoRA Fine-tuning}

For all training methods, we apply standard LoRA fine-tuning to all attention and MLP layers with $r=16$, $\alpha=32$, and dropout$=0.1$. We also use BF16 mixed precision and a cosine learning-rate schedule with warm-up steps. Other hyperparameters vary across models and training configurations and are selected using a small held-out set for each dataset. Details are provided in Table~\ref{tab:hparams}.

\subsection{Learning Rate Sensitivity}
\label{appx:lr-sensitivity}

\begin{figure}[ht]
\centering
\includegraphics[width=\columnwidth]{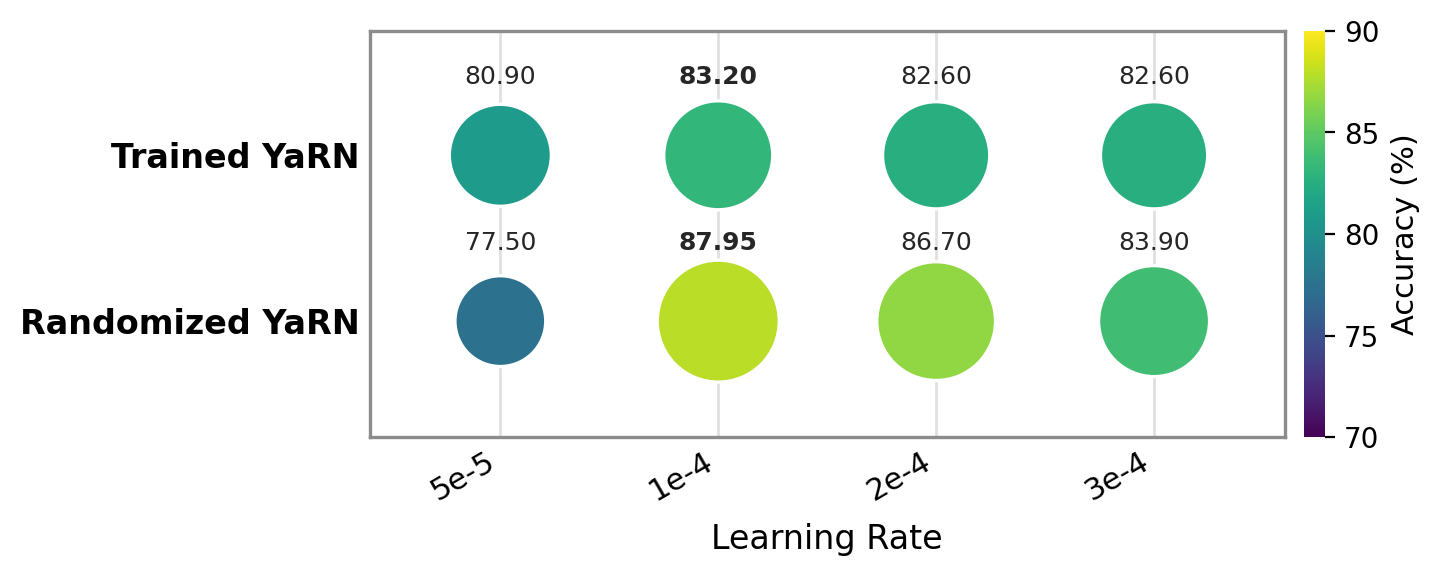}
\caption{Learning-rate sweep results on BABILong for Olmo3-7B-Instruct. OOD average is reported in the figure.}
\label{figure:lr-sweep}
\end{figure}

To investigate how sensitive our method is to hyperparameters, particularly learning rate, we present results of  a learning-rate sweep for trained YaRN and Randomized YaRN on Olmo3-7B-Instruct for BABILong, over $5e{-}5$, $1e{-}4$, $2e{-}4$, and $3e{-}4$, and training for 5 epochs. As shown in Figure~\ref{figure:lr-sweep}, both the baseline and Randomized YaRN peak at $1e{-}4$, which is the learning rate we report in the main results; here Randomized YaRN exceeds Trained YaRN by 4.8\%. Across different learning rates, Randomized YaRN is generally better than the baseline.

\section{Method Implementation}
\label{appx:baseline_impl}

\paragraph{YaRN}

As discussed in Section~\ref{sec:randomized-yarn}, the scale factor is an important hyperparameter for YaRN during both training and inference. Let $s$ denote the scale factor used during training and $s'$ the factor used during inference. When $s'=s$, inference extrapolation matches training-time extrapolation, allowing the context length during inference to extend to approximately $s \cdot L_{\mathrm{pre}}$. Surprisingly, we empirically observe that settings where $s<s'$ can perform well in certain cases, suggesting that YaRN can learn length generalization during training using a smaller extrapolation range and later generalize to a larger extrapolation range at test time (from $s \cdot L_{\mathrm{pre}}$ to $s' \cdot L_{\mathrm{pre}}$). Therefore, we treat $s$ and $s'$ as separate hyperparameters and tune them independently on held-out datasets. Detailed configurations for all methods, including baselines, are provided in Table~\ref{tab:method-config}.

\paragraph{RPE}

We consider two hyperparameters for RPE: the maximum extrapolation length $L_T$ and the length curriculum. Details for each method can be found in Table~\ref{tab:method-config}.

\paragraph{PoSE}

PoSE \citep{zhu2023pose} divides the original context window into $N$ chunks and assigns each a \emph{skipping bias} that shifts its position indices, inserting a gap between consecutive chunks. Following \citet{zhu2023pose} we use $N{=}2$, and a single fixed target length of 32K. As with RPE, the manipulation is applied only during training, with YaRN at inference. Details for each method can be found in Table~\ref{tab:method-config}.

\paragraph{Compute and Resources}
Olmo3-7B-Instruct training experiments were run on a single NVIDIA A100-SXM4 GPU with 80 GB memory, and evaluations were on 2 GPUs. Qwen2.5-7B-Instruct and Qwen3-14B training experiments were run on a single NVIDIA H200 GPU with 141GB Memory. Total compute across all reported training and evaluation runs is approximately 275 GPU-hours.

\section{Licenses}

We use the following publicly available datasets from prior works with open licenses.
\paragraph{MRCR} MRCR uses the MIT license and is available at \url{https://huggingface.co/datasets/openai/mrcr}. 

\paragraph{BABILong} BABILong uses Apache 2.0 license and data is available at: 
\url{https://github.com/booydar/babilong}.

\paragraph{LongBench v2} LongBench v2 uses the Apache 2.0 license and data is available at: 
\url{https://huggingface.co/datasets/zai-org/LongBench-v2}.

\end{document}